\documentclass[journal]{IEEEtran}
\usepackage{amsmath,amsfonts}
\usepackage{algorithmic}
\usepackage{algorithm}
\usepackage{array}
\usepackage[caption=false,font=normalsize,labelfont=sf,textfont=sf]{subfig}
\usepackage{textcomp}
\usepackage{stfloats}
\usepackage{url}
\usepackage{verbatim}
\usepackage{graphicx}
\usepackage{cite}
\usepackage{xcolor}
\usepackage{xspace}
\usepackage{booktabs}
\usepackage{multirow}
\usepackage{graphicx}

\newcommand{\ourpaper}{ThermoSplat\xspace}
\newcommand{\szq}[1]{{#1}}

\begin{document}

\title{ThermoSplat: Cross-Modal 3D Gaussian Splatting with Feature Modulation and Geometry Decoupling}

\author{Zhaoqi Su, Shihai Chen, Xinyan Lin, Liqin Huang, Zhipeng Su, and Xiaoqiang Lu
\thanks{Zhaoqi Su, Shihai Chen, Xinyan Lin, Liqin Huang, Zhipeng Su, and Xiaoqiang Lu are with the College of Physics and Information Engineering, Fuzhou University, Fuzhou 350108, China.}
\thanks{E-mail: \{suzhaoqi, 2501120132, 2501127081, hlq, szp01, luxiaoqiang\}@fzu.edu.cn}
\thanks{Corresponding author: Zhipeng Su, Xiaoqiang Lu}
}



\maketitle

\begin{abstract}
Multi-modal scene reconstruction integrating RGB and thermal infrared data is essential for robust environmental perception across diverse lighting and weather conditions. However, extending 3D Gaussian Splatting (3DGS) to multi-spectral scenarios remains challenging. Current approaches often struggle to fully leverage the complementary information of multi-modal data, typically relying on mechanisms that either tend to neglect cross-modal correlations or leverage shared representations that fail to adaptively handle the complex structural correlations and physical discrepancies between spectrums. To address these limitations, we propose \ourpaper, a novel framework that enables deep spectral-aware reconstruction through active feature modulation and adaptive geometry decoupling. First, we introduce a Spectrum-Aware Adaptive Modulation that dynamically conditions shared latent features on thermal structural priors, effectively guiding visible texture synthesis with reliable cross-modal geometric cues. Second, to accommodate modality-specific geometric inconsistencies, we propose a Modality-Adaptive Geometric Decoupling scheme that learns independent opacity offsets and executes an independent rasterization pass for the thermal branch. Additionally, a hybrid rendering pipeline is employed to integrate explicit Spherical Harmonics with implicit neural decoding, ensuring both semantic consistency and high-frequency detail preservation. Extensive experiments on the RGBT-Scenes dataset demonstrate that \ourpaper achieves state-of-the-art rendering quality across both visible and thermal spectrums.
\end{abstract}


\section{Introduction}
3D scene reconstruction has been widely used in the field of autonomous systems, remote sensing, surveillance, etc.
Traditional RGB-based reconstruction methods, while achieving high fidelity in most conditions, often suffer from performance degradation in challenging environments, e.g., low-light conditions, dense smoke, or darkness. To address these limitations, multi-modal reconstruction, especially those integrating RGB and thermal, has emerged as a critical research direction. Unlike RGB sensors, which depend on reflected light, thermal sensors capture long-wave infrared radiation emitted by objects, allowing for the acquisition of stable structural information and heat signatures that are inherently invariant to illumination changes. This provides a reliable reference for scene geometry under extreme conditions.

The evolution of neural implicit representations has inspired research in multi-modal 3D reconstruction. Previous studies~\cite{thermalnerf, thermonerf} extended the Neural Radiance Fields (NeRF)~\cite{nerf} framework to the infrared spectrum, demonstrating the potential to synthesize thermal views from multi-view observations. However, NeRF-based methods often suffer from high computational cost and slow inference speeds, limiting their abilities in real-time applications. Recently, the emergence of 3D Gaussian Splatting (3DGS)~\cite{3dgs} has enabled significantly faster training and rendering, with improved rendering quality. Building on this, several multi-spectral 3DGS frameworks have been proposed. 
Current state-of-the-art methods can be categorized into two paradigms: either explicitly decomposing 3DGS into modality-specific components to handle property disparities~\cite{thermalgaussian}, or integrating multi-spectral information into a unified latent space for MLP-based decoding~\cite{mssplatting, mmone}. However, achieving an optimal balance between shared geometry and modality-specific appearance remains non-trivial. The former paradigm often introduces increased model complexity and may face challenges in maintaining cross-modal spatial consistency, while the latter tends to have a limited capacity to precisely model the inherent physical discrepancies and structural variations present across different spectrums.

To bridge these gaps, we present \ourpaper, a novel cross-modal 3DGS framework that enables deep spectral-aware reconstruction through active feature modulation and adaptive geometry decoupling. Unlike existing methods that either rely on explicit decomposition or unified latent representation, \ourpaper introduces a Spectrum-Aware Adaptive Modulation. While drawing inspiration from the mathematical flexibility of linear modulation~\cite{film}, this module dynamically conditions the shared latent representation on thermal structural priors, enabling the model to actively leverage structural infrared features to guide visible texture synthesis. Furthermore, to accommodate the inherent physical discrepancies across different spectral bands, we propose a modality-adaptive geometric decoupling scheme, which allows for independent geometric adjustments in the infrared spectrum, effectively resolving artifacts in regions where transparency or reflectivity varies. Finally, to overcome the detail-loss inherent in pure feature-based decoding, we employ a hybrid rendering pipeline, which integrates explicit Spherical Harmonics (SH) with implicit decoding, achieving high-frequency RGB details while maintaining consistent semantic information across modalities.

Experimental results demonstrate that \ourpaper achieves state-of-the-art rendering quality across both visible and thermal spectrums. The main contributions of this work are summarized as follows:

\begin{itemize}
    \item \textbf{Spectrum-Aware Adaptive Modulation:} We design a Spectrum-Aware Adaptive Modulation framework to establish deep feature dependencies. By utilizing structural priors to modulate shared latent features, our method enhances texture recovery and cross-modal alignment.
    
    \item \textbf{Modality-Adaptive Geometric Decoupling:} We introduce a learnable thermal opacity offset and execute an independent rasterization pass that decouples geometric representations between visible and infrared spectrums. This mechanism effectively resolves depth and occlusion misalignments caused by modality-inconsistent physical properties.
    
    \item \textbf{Hybrid Explicit-Implicit Rendering:} We propose a hybrid rendering pipeline that integrates explicit Spherical Harmonics (SH) with feature-modulated neural decoding. This architecture preserves high-frequency RGB details while maintaining consistent low-frequency semantic information across different modalities.
\end{itemize}

\section{Related Work}

\subsection{3D Neural Scene Representation}

Recent methods in 3D neural scene representation have shifted from implicit to explicit methods. The implicit NeRF-based methods~\cite{nerf, mipnerf, instantngp, nerfstudio} represent the scene as a continuous function in 3D space formulated by a shallow network like MLP, achieving view-dependent and photo-realistic scene rendering results. However, these methods suffer from time-consuming training and rendering, limiting their practical use in real-time applications. To address this, 3D Gaussian Splatting (3DGS)-based methods~\cite{3dgs, 3dgssurvey, mipsplatting, scaffoldgs} propose an explicit scene representation paradigm, which leverages 3D Gaussian primitives for explicitly representing the geometric and texture information of the scene, enabling high-fidelity and real-time rendering through a differentiable tile-based rasterization pipeline. Some studies~\cite{feature3dgs, efficientfeature3dgs, featuresplatting} leverage the idea of both feature-based decoding in NeRF and 3DGS representations to augment the representation capability by distilling high-dimensional latent features into each Gaussian primitive. By integrating these latent features with lightweight decoders, these methods can bypass the limitations of traditional Spherical Harmonics (SH), enabling more complex attribute modeling and cross-modal information interaction.

\subsection{Neural Thermal and RGBT Scene Reconstruction}
Compared to visible light, thermal infrared signals possess distinct physical properties, such as being insensitive to lighting conditions and capable of reflecting the heat distribution of objects. Early attempts mostly extend NeRF-based representations for representing different modalities in a compact manner~\cite{thermalnerf, leveragingthermalnerf, thermal-nerf, exploringthermalnerf, thermonerf}. However, due to the volume rendering process in NeRF~\cite{nerf} and its reliance on dense sampling, these implicit methods often face challenges in precisely modeling the high-frequency details. In recent years, the emergence of 3DGS has shifted the focus toward explicit Gaussian-based RGBT (RGB + Thermal) scene modeling. ThermalGaussian~\cite{thermalgaussian} pioneered the extension of 3DGS to the RGBT scene, which optimizes the thermal Gaussian by fine-tuning the pretrained RGB Gaussians and incorporates thermal priors for better scene modeling. Also, it releases the RGBT-Scenes dataset to facilitate benchmarking for multi-modal reconstruction tasks. MS-Splatting~\cite{mssplatting} formulates the multi-spectral 3D scene using a unified latent space for decoding both RGB and other spectral channels, which is also applied to agricultural NDVI tasks. MS-Splattingv2~\cite{mssplattingvmv} uses the optimized joint strategy with RGB initialization to improve rendering quality. MMOne~\cite{mmone} introduces a unified framework that represents multiple modalities, such as RGB, thermal, and language, within a single scene, which designs a multimodal decomposition mechanism for better learning properties of different modalities. 
Ma et al.~\cite{beyonddarkness} decomposes appearance into reflectance and thermal radiance, leveraging the thermal modality as a stable geometric prior to rectify distorted surfaces in low-light RGB inputs. 
Beyond general multimodal representation, several studies focus on reconstructing thermal infrared signals to tackle ill-posed problems or extreme environmental constraints. 
Some studies~\cite{thermal3d-gs, ntrgaussian} inject physics-based temperature or thermodynamics constraints into thermal 3DGS modeling.
Veta-GS~\cite{vetags} introduces a view-dependent deformation field to capture the subtle thermal variations caused by emissivity and transmission effects, effectively reducing artifacts in infrared novel-view synthesis. 
Others extend the RGBT modeling into more-spectral or hyperspectral scenarios~\cite{spectralgaussians, hypergs}.
Despite these advances, existing RGBT frameworks either treat different modalities as independent signals with limited feature interaction, or rely on a shared representation that tends to overlook modality-specific physical discrepancies. These limitations motivate us to explore a more flexible modulation and decoupled modeling for RGBT scene reconstruction.

\section{Method}

\subsection{Overview}
\begin{figure*}[!ht]
\centering
\includegraphics[width=\linewidth]{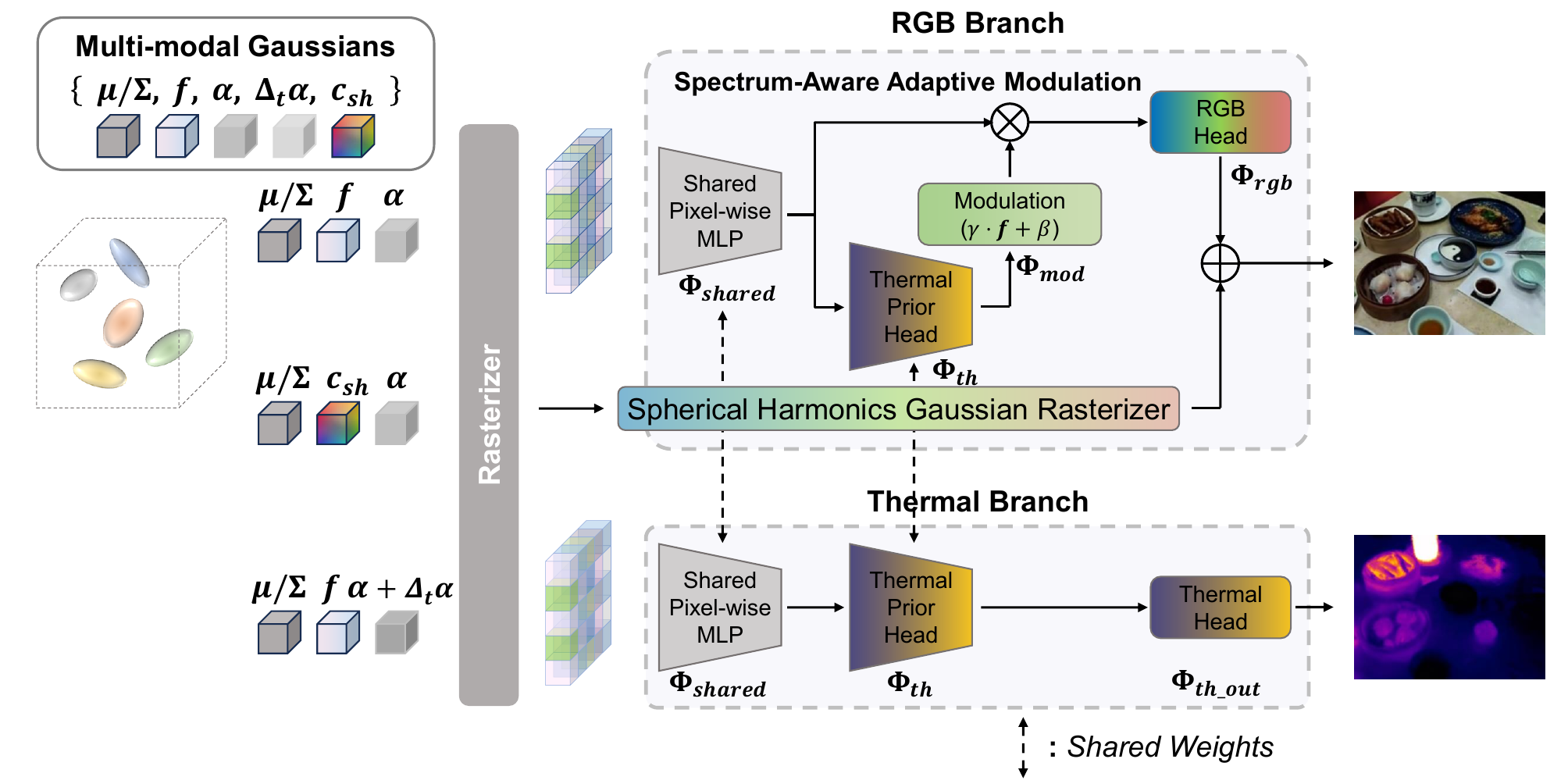}
\caption{\szq{Overview of the proposed \ourpaper framework. Given multi-spectral inputs, our method optimizes 3D Gaussian primitives with decoupled properties. (a) Spectrum-Aware Adaptive Modulation dynamically conditions shared latent features on thermal structural priors to guide visible texture synthesis. (b) Modality-Adaptive Geometric Decoupling resolves geometric inconsistencies between visible and infrared spectrums. (c) The Hybrid Rendering pipeline integrates explicit Spherical Harmonics (SH) with implicit neural decoding, ensuring high-frequency detail preservation and cross-modal semantic consistency.}}
\label{fig:pipeline}
\end{figure*}

The overall architecture of \ourpaper is designed to achieve high-fidelity multi-modal scene reconstruction by addressing spectral-varying properties. As illustrated in Fig.~\ref{fig:pipeline}, we represent the scene using multi-modal feature-enhanced 3DGS~\cite{3dgs}.
The pipeline first performs active feature interaction via a \textbf{Spectrum-Aware Adaptive Modulation} on the rasterized latent representations, which utilizes thermal structural priors to guide visible texture synthesis. To account for geometric inconsistencies across spectrums, we introduce a \textbf{modality-adaptive geometric decoupling} scheme, which uses the learnable offset $\Delta_t\alpha$ and executes an independent rasterization pass to accommodate modality-specific geometries. Finally, a \textbf{hybrid rendering} strategy is employed to combine explicit Spherical Harmonics (SH) with implicit feature-decoded outputs for preserving high-frequency details and view-dependent effects in the visible spectrum.

The remainder of this section provides a detailed formalization of our framework. We first briefly introduce 3D Gaussian Splatting and feature-based rasterization in Section~\ref{sec:preliminaries}. In Section~\ref{sec:modulation}, we describe the cross-modal feature modulation mechanism, which enables active spectral interaction. Section~\ref{sec:rendering} presents the Multi-spectral Hybrid Rendering pipeline, where we first detail the modality-adaptive geometric decoupling for modal-specific geometries, followed by the hybrid rendering strategy for RGB synthesis to preserve high-frequency details. 
\szq{Finally, Section~\ref{sec:loss} elaborates the loss functions used in our pipeline.}

\subsection{Preliminaries} \label{sec:preliminaries}
3D Gaussian Splatting (3DGS)~\cite{3dgs} represents a 3D scene as a collection of $N$ Gaussian primitives. Each Gaussian is characterized by its center position $\mu \in \mathbb{R}^3$, an anisotropic covariance $\Sigma = RSS^T R^T$, and an opacity value $\alpha$. The influence of a Gaussian at a 3D point $\boldsymbol{x}'$ is defined as:
\begin{equation}
    G(\boldsymbol{x}'; \boldsymbol{\mu}_i, \Sigma_i) = e^{-\frac{1}{2}(\boldsymbol{x}'-\boldsymbol{\mu}_i)^T\Sigma_i^{-1}(\boldsymbol{x}'-\boldsymbol{\mu}_i)},
\end{equation}
where $\boldsymbol{x}'$ denotes the 3D point in the camera coordinate system. Unlike traditional 3DGS that directly optimizes Spherical Harmonic (SH) coefficients for color, we follow a feature-based splatting paradigm~\cite{mssplatting} where each Gaussian carries a multi-dimensional latent feature $f \in \mathbb{R}^d$. This feature serves as a unified latent representation that can be subsequently decoded into modality-specific signals (e.g., RGB or thermal) via neural networks.

The rendering process follows the point-based $\alpha$-blending model. For a specific pixel, the attributes of projected 2D Gaussians are sorted by depth and blended to compute the aggregated pixel value:
\begin{equation}
\mathcal{A} = \sum_{i \in \mathcal{N}} \mathbf{a}_i \alpha_i \prod_{j=1}^{i-1} (1 - \alpha_j),
\end{equation}
where $\mathbf{a}_i$ denotes the generic attribute of the $i$-th Gaussian (such as latent feature $f_i$ or explicit color attributes) and $\alpha_i$ is the opacity of the Gaussian at that pixel. 

In this work, we leverage this differentiable rasterization to bridge different modalities. By decoupling the feature decoding from the geometric projection, we can perform complex cross-modal modulations in the latent space before generating the final visible and thermal images.

\subsection{Cross-Modal Feature Modulation} \label{sec:modulation}
To address the spectral gap between visible and infrared modalities, we propose a cross-modal modulation mechanism. Instead of treating visible and thermal signals as independent entities, our framework leverages structural priors inherent in the thermal spectrum to guide the synthesis of visible textures. As shown in Fig.~\ref{fig:pipeline}, the proposed Spectrum-Aware Adaptive Modulation integrates feature extraction and conditioning in a unified neural architecture.

\textbf{Shared Latent Encoding.}
The process begins with the rendered feature map $\mathcal{A}_f \in \mathbb{R}^{H \times W \times d}$ by rasterizing the per-Gaussian feature $f_i$ through the 3DGS rendering pipeline~\cite{3dgs}. To extract high-level semantic information, we first pass $\mathcal{A}_f$ through a shared encoder $\Phi_{shared}$ consisting of multiple pixel-wise linear layers with SiLU activations~\cite{silu}:
\begin{equation}
    h = \Phi_{shared}(\mathcal{A}_f),
\end{equation}
where $h$ represents the intermediate feature representation that serves as the common basis for both modalities.

\textbf{Spectrum-Aware Adaptive Modulation.}
Distinct from directly applying MLP layers for different spectrums, we introduce a \textit{Thermal Prior Head} $\Phi_{th}$ and a \textit{Spectrum-Aware Adaptive Modulation}~\cite{film} layer. Unlike generic conditioning methods, our modulation parameters $(\gamma, \beta)$ are dynamically derived from a dedicated \textit{Thermal Prior Head} $\Phi_{th}$. Crucially, as $h_{th}=\Phi_{th}(h)$ is directly supervised by the thermal rendering task through the subsequent decoding stage (Eq.~\ref{eq:decode}), the modulation process is implicitly physics-aware, ensuring that the distilled structural priors from the infrared domain actively guide the visible feature synthesis. The feature $h_{th}$ is subsequently mapped to a set of modulation parameters $(\gamma, \beta)$ through a linear transformation:
\begin{equation}
    [\gamma, \beta] = \Phi_{mod}(h_{th}).
\end{equation}
\szq{Here, $\gamma$ and $\beta$ are vectors of the same dimension as $h_th$, enabling spatially‑varying feature modulation.}

By treating the infrared information as a conditioning signal, we apply Spectrum-Aware Adaptive Modulation to the shared representation $h$:
\begin{equation}
    h_{mod} = \gamma \odot h + \beta,
\end{equation}
where $\odot$ denotes element-wise multiplication. This operation dynamically scales and shifts the latent features based on the thermal structural prior, 
\szq{effectively performing a feature-level cross-modal modulation that selectively amplifies or suppresses feature channels according to the thermal context.}

\begin{figure*}[!t]
\centering
\includegraphics[width=\linewidth]{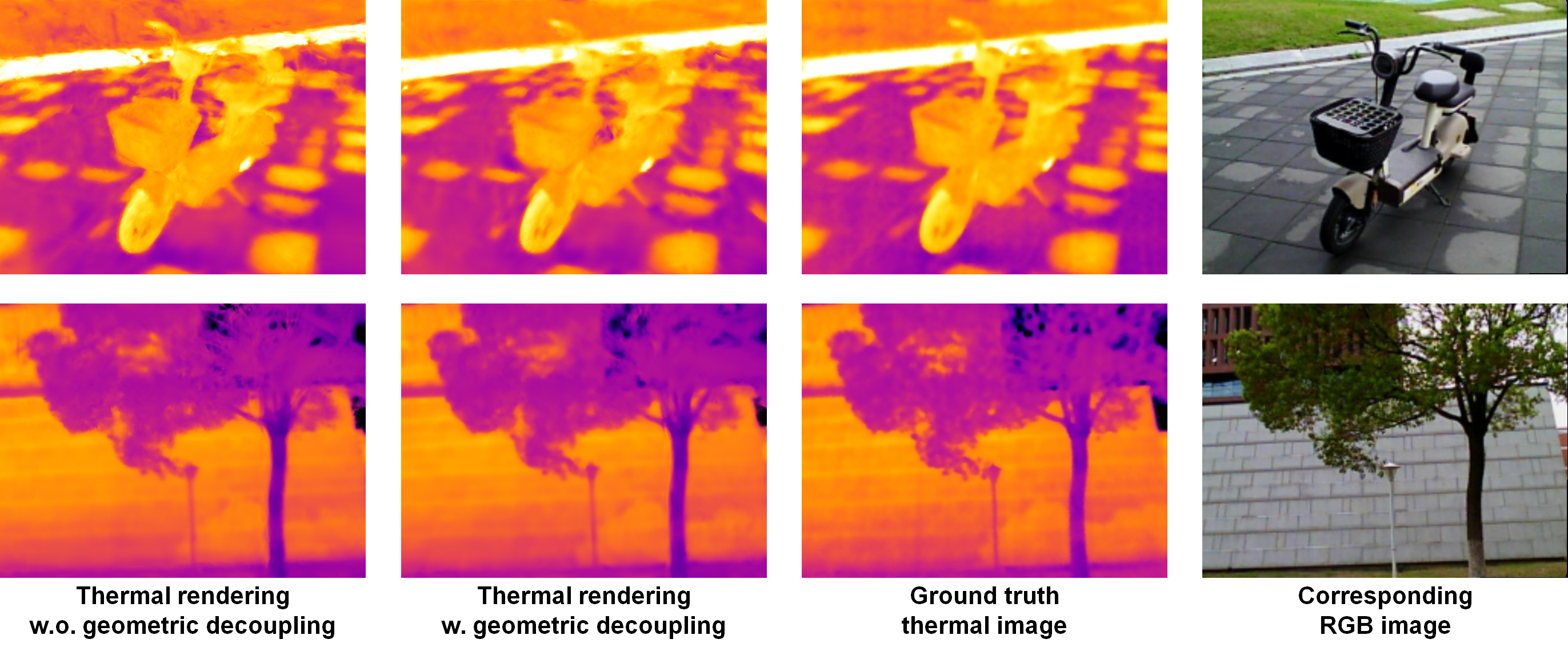}
\caption{Thermal rendering results with and without geometric decoupling. The thermal rendering results without geometric decoupling may inherit sharp textures and high-frequency noise from the visible spectrum.}
\label{fig:cmp_thop}
\end{figure*}

\textbf{Modality-Specific Decoding.}
Finally, the modulated feature $h_{mod}$ and the thermal feature $h_{th}$ are decoded into their respective spectral domains:
\begin{equation}
\label{eq:decode}
\begin{aligned}
    \mathcal{C}_{implicit}^{rgb} &= \text{Sigmoid}(\Phi_{rgb}(h_{mod})), \\
    \mathcal{C}^{thermal} &= \text{Sigmoid}(\Phi_{th\_out}(h_{th})),
\end{aligned}
\end{equation}
where $\mathcal{C}_{implicit}^{rgb}$ provides the base color component for the subsequent hybrid rendering stage. Notably, the thermal-specific feature $h_{th}$ serves a dual purpose: it acts as the source for the proposed modulation parameter generation while simultaneously being decoded into the infrared signal $\mathcal{C}^{thermal}$. This hierarchical modulation ensures that the synthesis of visible images is physically constrained by the cross-modal structural consistency.

\subsection{Multi-spectral Hybrid Rendering} \label{sec:rendering}
Based on the modulated cross-modal features, we develop a dual-branch rendering pipeline to synthesize images in both visible and thermal spectrums. This pipeline addresses the geometric inconsistencies and texture fidelity requirements unique to each modality.

\textbf{Modality-Adaptive Geometric Decoupling}
Typical multi-modal Gaussian representations assume a shared geometry across all spectrums. However, physical properties such as transparency and reflectivity vary significantly between visible and infrared bands. To accommodate these discrepancies, we introduce a modality-adaptive geometric decoupling scheme. 

For the thermal rendering branch, we define a modality-specific opacity $\alpha_{t,i}$ for each Gaussian $i$ by adding a learnable offset $\Delta_t\alpha_i$ to the base opacity $\alpha_i$:
\begin{equation}
    \alpha_{t,i} = \text{Sigmoid}(\text{Logit}(\alpha_i) + \Delta_t\alpha_i),
\end{equation}
where $\Delta_t\alpha_i$ captures the fine-grained geometric deviations. Consequently, the thermal representation is generated via an independent rasterization pass:
\begin{equation}
    \mathcal{A}_{f(t)} = \text{Rasterize}(\mu, \Sigma, \alpha_{t}, f),
\end{equation}
where $\mathcal{A}_{f(t)} \in \mathbb{R}^{H \times W \times d}$ represents the thermal-specific feature map generated using the decoupled opacity $\alpha_t$. The final thermal image $\mathcal{C}^{thermal}$ is then decoded from $\mathcal{A}_{f(t)}$ 
\szq{via the shared MLP and the thermal heads}
discussed in Section~\ref{sec:modulation}. This independent pass ensures that occlusions and structural boundaries in the thermal image remain physically consistent with infrared sensors.

As shown in Fig.~\ref{fig:cmp_thop}, without the geometric decoupling mechanism, the thermal branch tends to inherit redundant high-frequency textures from the visible spectrum that do not exist in the infrared domain. Our proposed decoupling module effectively filters out these cross-modal artifacts, ensuring that the thermal rendering preserves its natural smoothness while accurately representing its own structural boundaries.

\textbf{Hybrid RGB Synthesis.}
While the thermal branch focuses on geometric consistency, the RGB branch requires high-frequency view-dependent details. We propose a hybrid strategy that bridges explicit Gaussian Splatting with implicit neural decoding.

Specifically, the final RGB color $\mathbf{C}^{rgb}$ is formulated as the summation of two components:
\begin{equation}
    \mathbf{C}^{rgb} = \mathcal{R}_{sh}(\mu, \Sigma, \alpha, \mathbf{c}_{sh}) \oplus \mathcal{C}_{implicit}^{rgb},
\end{equation}
where $\mathcal{R}_{sh}$ denotes the explicit color rendered via standard Spherical Harmonic (SH) coefficients $\mathbf{c}_{sh}$, capturing view-dependent specular effects. The second term, $\mathcal{C}_{implicit}^{rgb}$, is the implicit component decoded from the modulated latent features $h_{mod}$, providing multi-modal consistent textures. By combining these two components, our hybrid rendering scheme effectively preserves the high-frequency view-dependent properties of explicit rasterization, while simultaneously enriching the visible textures with the structural intelligence of neural-modulated latent features.

\subsection{Loss Functions} \label{sec:loss}
The training objective of \ourpaper is to optimize the multi-modal Gaussian representation and the neural modulation networks through a composite loss function $\mathcal{L}$. This objective ensures that the synthesized visible and thermal images adhere to the ground truth in terms of both pixel intensity and structural topology.

\textbf{Spectral Reconstruction Loss}
For both the visible and infrared modalities, we employ a combination of $\ell_1$ loss and Structural Similarity (SSIM) to supervise the final rendered images against the corresponding ground-truth images $\mathbf{I}_m$:
\begin{equation}
\label{eq:loss_rec}
    \mathcal{L}_{rec}^m = (1 - \lambda_{s}) \|\mathbf{I}_m - \mathbf{C}^m\|_1 + \lambda_{s} (1 - \text{SSIM}(\mathbf{I}_m, \mathbf{C}^m)),
\end{equation}
where $m \in \{rgb, thermal\}$ denotes the spectral modality, and $\mathbf{C}^{m}$ are the corresponding output images in our pipeline. 

\begin{figure}[!t]
\centering
\includegraphics[width=\linewidth]{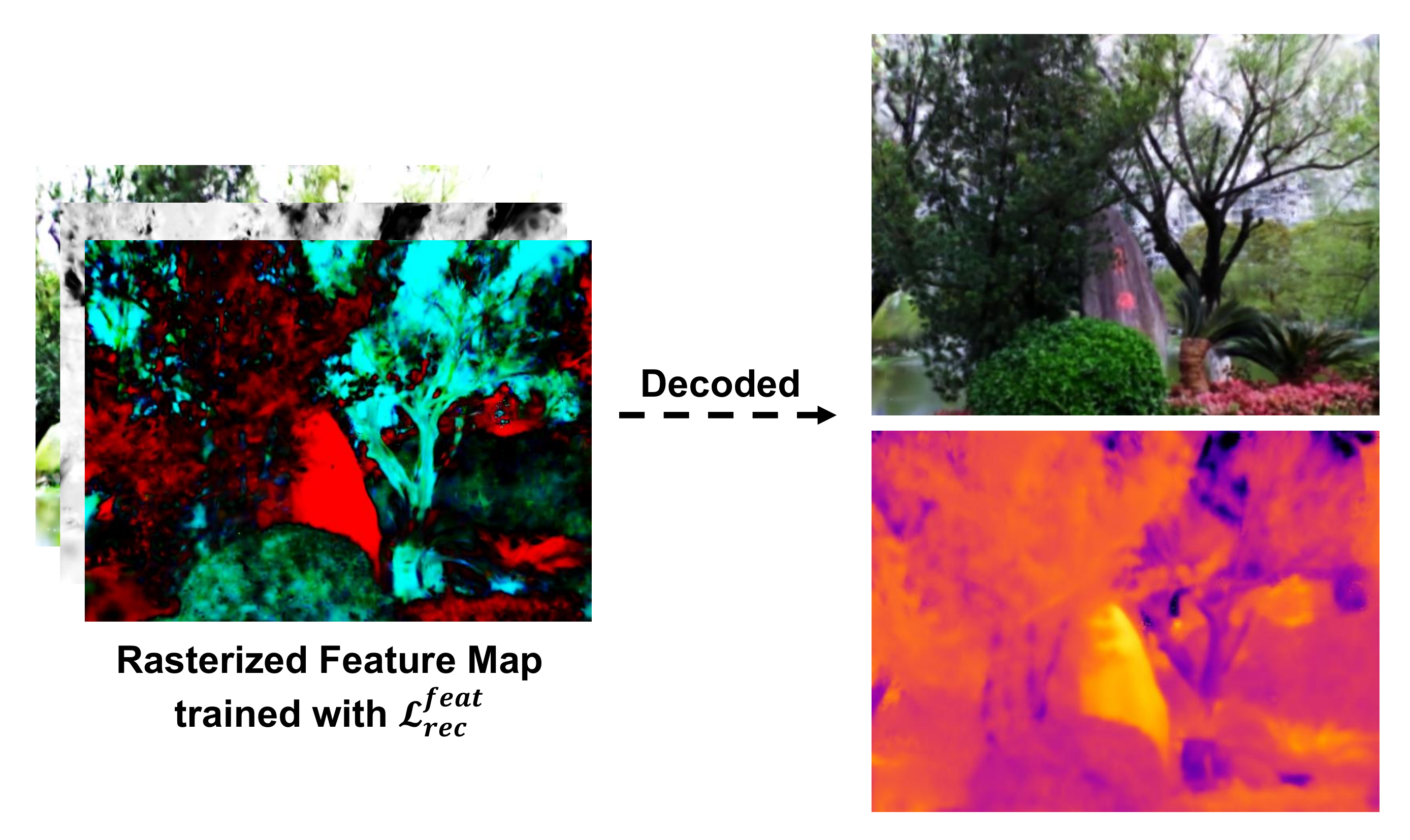}
\caption{Feature level reconstruction loss on the rasterized feature maps. Left: rendered feature map, right: reconstructed RGB-thermal scene. Note that $\mathcal{A}_{f}$ and $\mathcal{A}_{f(t)}$ are only different in the opacity used in rasterization.}
\label{fig:loss_feat}
\end{figure}

To provide structural guidance during the intermediate stages, we enforce consistency on the rasterized feature maps $\mathcal{A}_{f}$ and $\mathcal{A}_{f(t)}$ by slicing specific channels corresponding to physical properties.
As shown in Fig.~\ref{fig:loss_feat}, for the visible branch, we constrain the first three channels of $\mathcal{A}_{f}$ to match the RGB appearance. In parallel, for the thermal branch, we supervise the subsequent latent channel (index 3) of $\mathcal{A}_{f(t)}$ using the transformed thermal map derived from the Ironbow colormap protocol. As this transformed map effectively serves as a proxy for physical temperature and thermal intensity, this constraint encourages the model to learn a compact and structural representation. By applying these latent constraints, we ensure the latent space captures the fundamental visual and thermal distribution before it is decoded.
The feature-level reconstruction loss is thus formulated as:
\begin{equation}
    \mathcal{L}_{rec}^{feat} = \mathcal{L}(\mathcal{A}_{f}[:3], \mathbf{I}_{rgb}) + \eta \cdot \mathcal{L}(\mathcal{A}_{f(t)}[3], \mathbf{I}_{th}^{trans}),
\end{equation}
where $\mathbf{I}_{th}^{trans}$ represents the temperature-correlated intensity map, $\mathcal{L}$ denotes the composite $\ell_1$ and SSIM loss function as defined in Eq.~\ref{eq:loss_rec}. \szq{Unlike pseudo‑color maps, which are non‑linear and colormap‑specific, the Ironbow‑transformed intensity map preserves a monotonic relation with thermal radiation, providing a physically meaningful and linear proxy for latent space regularization.}

\textbf{Thermal Spatial Regularization}
Due to the high-contrast and often sparse nature of infrared signals, we introduce a spatial smoothness constraint on the predicted thermal image:
\begin{equation}
    \mathcal{L}_{smooth} = \sum_{p \in \Omega} |\nabla \mathbf{C}^{thermal}(p)|,
\end{equation}
where $\nabla$ denotes the spatial gradient operator at pixel $p$. This term enforces the smooth structural characteristics of the thermal output.

\textbf{Total Objective}
The final training objective is a weighted summation of the aforementioned reconstruction and regularization terms:
\begin{equation}
    \mathcal{L} = \mathcal{L}_{rec} + \lambda_{rf} \mathcal{L}_{rec}^{feat} +  \lambda_{sm} \mathcal{L}_{smooth},
\end{equation}
where the image-level reconstruction loss is defined as $\mathcal{L}_{rec} = \mathcal{L}_{rec}^{rgb} + \mathcal{L}_{rec}^{thermal}$, $\lambda_{rf}$ and $\lambda_{sm}$ are hyper-parameters balancing feature terms and smooth terms. By optimizing this joint objective, our framework ensures that the synthesized modalities satisfy both pixel-level accuracy and the inherent structural characteristics of thermal radiation.

\section{Experiments}

\subsection{Implementation details.}

We evaluate our model on the RGBT-Scenes dataset, which comprises over 1,000 calibrated RGB-thermal pairs across ten indoor and outdoor scenes under diverse environmental and lighting conditions. To demonstrate the effectiveness of our approach, we compare our model with state-of-the-art methods, including MMOne~\cite{mmone}, MS-Splattingv2~\cite{mssplattingvmv}
and ThermalGaussian~\cite{thermalgaussian}. We also compare our method with the 3DGS~\cite{3dgs} baseline trained on both modalities separately. We use Peak Signal-to-Noise
Ratio (PSNR), Structural Similarity Index (SSIM), and
Learned Perceptual Image Patch Similarity (LPIPS)~\cite{lpips} to evaluate the rendering quality of both visible and thermal modalities.

Our framework is implemented on PyTorch and trained with an NVIDIA 3090 GPU. We train our pipeline for 30K iterations, which is the same as the setting used in 3DGS~\cite{3dgs}, MMOne~\cite{mmone} and ThermalGaussian~\cite{thermalgaussian}. For 
MS-Splattingv2~\cite{mssplattingvmv}, we follow the training strategy proposed in the paper and train them with 
120K iterations.
\szq{Specifically, each scene is trained for approximately 30 minutes on a single NVIDIA 3090 GPU. Following the original RGBT-Scenes dataset protocol, we adopt an 8:1 train/test split for each scene. The hyperparameters, including the feature dimension $d=8$ and loss weights $\lambda_s=0.2$, $\eta=0.5$, $\lambda_{rf}=1$, $\lambda_{sm}=0.3$, are chosen as the optimal configuration based on our ablation experiments.}

\begin{figure*}[!t]
\centering
\includegraphics[width=\linewidth]{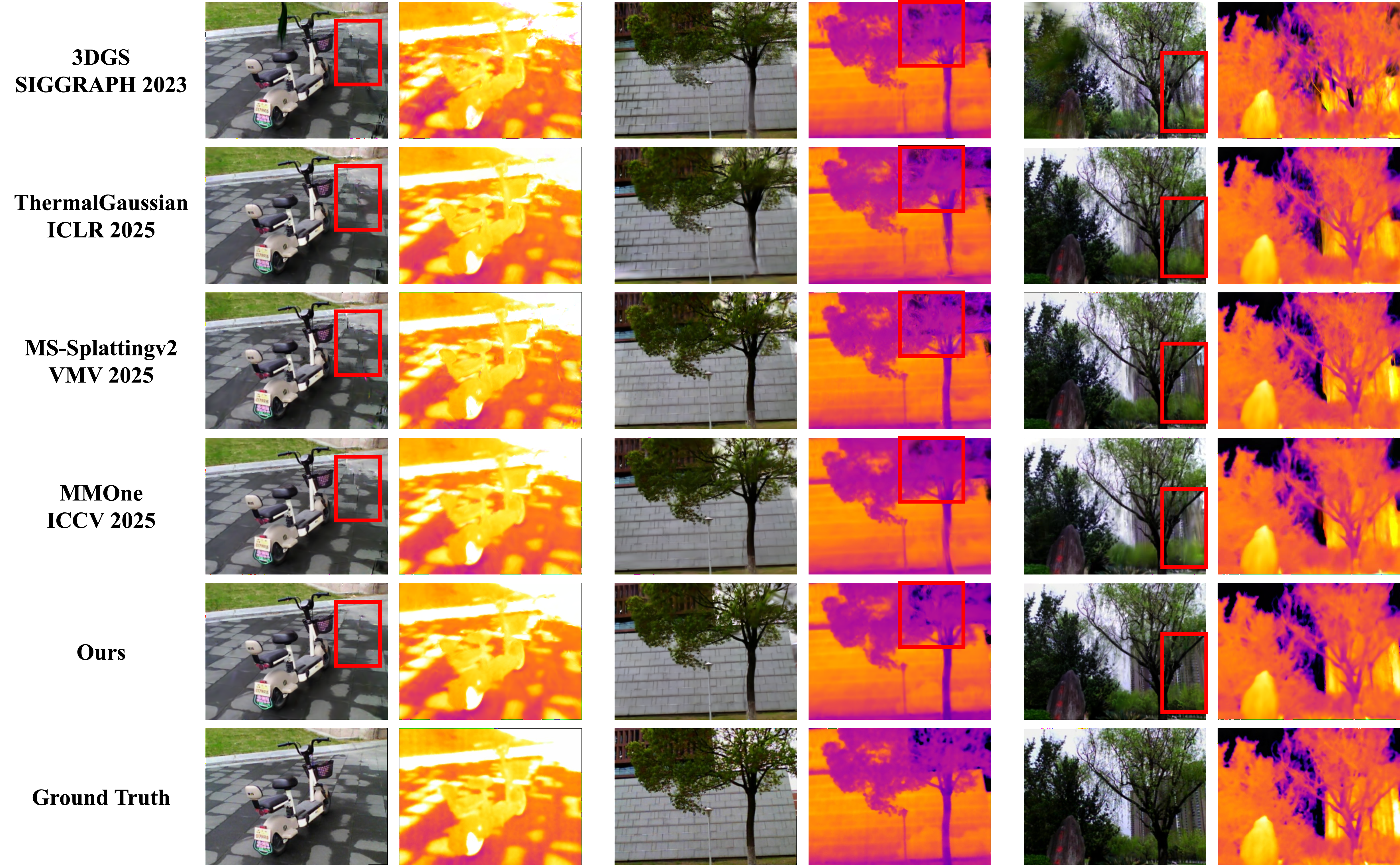}
\caption{\szq{Qualitative comparison of novel view synthesis results on the RGBT-Scenes dataset. We compare \ourpaper against state-of-the-art multi-spectral reconstruction methods ThermalGaussian~\cite{thermalgaussian}, MS-Splattingv2~\cite{mssplattingvmv}, MMOne~\cite{mmone}, and the 3DGS baseline~\cite{3dgs}. Our method generates more accurate rendering results and structural details.}}
\label{fig:results}
\end{figure*}

\subsection{Results and Comparisons}
We evaluate the novel view synthesis performance on the test set to validate the rendering quality of our method against other state-of-the-art methods. As illustrated in Fig.~\ref{fig:results}, our method produces results with finer texture details and fewer visual artifacts compared to existing approaches and the 3DGS baseline. Specifically, our model performs better in recovering complex structures that are often blurred or misaligned in the baseline reconstructions, especially on the RGB branch, which is attributed to the proposed cross-modality modulation that effectively leverages structural priors of the scene.

As shown in Tab.~\ref{tab:results}, our method achieves superior performance compared to state-of-the-art baselines across most scenes. Specifically, our model attains the highest average scores in all three metrics for both RGB and thermal modalities. Notably, on the average PSNR, our method outperforms the second-best competitor (MMOne~\cite{mmone}) by $0.34$ dB in the RGB spectrum and $0.19$ dB in the thermal spectrum. The consistent improvement in SSIM and LPIPS further demonstrates our model's capability to reconstruct fine-grained structural details and maintain perceptual fidelity.

While MMOne~\cite{mmone} and MS-Splattingv2~\cite{mssplattingvmv} show competitive results in specific scenes (e.g., Dim and Trk), our method demonstrates more robust generalization across diverse environments. These results validate that our modality modulation and geometric decoupling strategy successfully resolves the discrepancies between modalities without compromising the reconstruction quality of the individual branches.

\begin{table*}[!ht]
\centering
\caption{Quantitative evaluation of RGB and thermal (T) rendering results.}
\label{tab:results}
\resizebox{\linewidth}{!}{
\begin{tabular}{lllccccccccccc}
\toprule
M & Metric & Method & Dim & DS & Ebk & RB & Trk & RK & Bldg & II & Pt & LS & Avg. \\
\midrule
    RGB & PSNR $\uparrow$ & 3DGS            & $23.27$ & $21.18$ & $26.17$ & $28.23$ & $22.45$ & $20.74$ & $21.80$ & $24.40$ & $25.65$ & $20.18$ & $23.41$ \\
        &                  & ThermalGaussian & $24.38$ & $21.76$ & $26.85$ & $28.12$ & $24.17$ & $23.14$ & $\textbf{24.19}$ & $24.55$ & $25.48$ & $21.71$ & $24.44$ \\
        &                  & MS-Splattingv2  & $24.06$ & $21.18$ & $26.87$ & $28.12$ & $\textbf{24.54}$ & $23.42$ & $23.90$ & $23.77$ & $\underline{26.20}$ & $\underline{22.05}$ & $24.41$ \\
        &                  & MMOne           & $\textbf{24.65}$ & $\underline{22.05}$ & $\textbf{27.43}$ & $\textbf{29.03}$ & $23.96$ & $\underline{24.12}$ & $\underline{24.16}$ & $\underline{25.65}$ & $26.01$ & $21.81$ & $\underline{24.89}$ \\
        &                  & \textbf{Ours}   & $\underline{24.59}$ & $\textbf{22.12}$ & $\underline{27.21}$ & $\underline{28.96}$ & $\underline{24.31}$ & $\textbf{24.20}$ & $24.14$ & $\textbf{25.98}$ & $\textbf{26.48}$ & $\textbf{24.31}$ & $\textbf{25.23}$ \\
    \cmidrule(lr){3-14}
     & SSIM $\uparrow$ & 3DGS            & $0.842$ & $0.771$ & $0.902$ & $0.917$ & $0.810$ & $0.765$ & $0.827$ & $0.875$ & $0.867$ & $0.688$ & $0.826$ \\
        &                  & ThermalGaussian & $0.858$ & $0.797$ & $0.905$ & $0.920$ & $0.840$ & $0.822$ & $0.849$ & $0.884$ & $0.855$ & $0.739$ & $0.847$ \\
        &                  & MS-Splattingv2  & $0.859$ & $0.788$ & $0.914$ & $\underline{0.922}$ & $\textbf{0.859}$ & $0.827$ & $\underline{0.855}$ & $0.877$ & $\underline{0.878}$ & $\underline{0.739}$ & $0.852$ \\
        &                  & MMOne           & $\underline{0.862}$ & $\underline{0.810}$ & $\underline{0.918}$ & $0.916$ & $0.845$ & $\textbf{0.842}$ & $0.847$ & $\underline{0.897}$ & $0.876$ & $0.727$ & $\underline{0.854}$ \\
        &                  & \textbf{Ours}   & $\textbf{0.872}$ & $\textbf{0.818}$ & $\textbf{0.934}$ & $\textbf{0.941}$ & $\underline{0.859}$ & $\underline{0.841}$ & $\textbf{0.858}$ & $\textbf{0.911}$ & $\textbf{0.886}$ & $\textbf{0.788}$ & $\textbf{0.871}$ \\
    \cmidrule(lr){3-14}
     & LPIPS $\downarrow$ & 3DGS            & $0.199$ & $0.271$ & $0.169$ & $\underline{0.197}$ & $0.244$ & $0.220$ & $0.183$ & $0.193$ & $0.177$ & $0.289$ & $0.214$ \\
        &                  & ThermalGaussian & $0.194$ & $0.253$ & $0.169$ & $0.199$ & $0.211$ & $0.184$ & $0.170$ & $0.186$ & $0.195$ & $0.268$ & $0.203$ \\
        &                  & MS-Splattingv2  & $\textbf{0.150}$ & $\underline{0.224}$ & $\underline{0.145}$ & $0.197$ & $\underline{0.170}$ & $\underline{0.141}$ & $\underline{0.145}$ & $\underline{0.161}$ & $\textbf{0.132}$ & $\underline{0.211}$ & $\underline{0.168}$ \\
        &                  & MMOne           & $0.203$ & $0.254$ & $0.160$ & $0.235$ & $0.226$ & $0.178$ & $0.184$ & $0.183$ & $0.178$ & $0.291$ & $0.209$ \\
        &                  & \textbf{Ours}   & $\underline{0.155}$ & $\textbf{0.204}$ & $\textbf{0.121}$ & $\textbf{0.164}$ & $\textbf{0.166}$ & $\textbf{0.130}$ & $\textbf{0.131}$ & $\textbf{0.138}$ & $\underline{0.136}$ & $\textbf{0.180}$ & $\textbf{0.153}$ \\
    \midrule
    T & PSNR $\uparrow$ & 3DGS            & $25.99$ & $18.71$ & $20.61$ & $26.55$ & $25.30$ & $26.45$ & $26.83$ & $29.69$ & $24.09$ & $18.48$ & $24.27$ \\
        &                  & ThermalGaussian & $\underline{26.46}$ & $\textbf{22.28}$ & $23.31$ & $\underline{27.17}$ & $25.88$ & $26.33$ & $26.72$ & $29.86$ & $26.16$ & $22.27$ & $25.64$ \\
        &                  & MS-Splattingv2  & $26.06$ & $21.43$ & $\underline{23.32}$ & $25.44$ & $\underline{26.08}$ & $27.24$ & $26.89$ & $\underline{29.98}$ & $\textbf{27.01}$ & $\underline{22.64}$ & $25.61$ \\
        &                  & MMOne           & $\textbf{26.90}$ & $\underline{21.81}$ & $\textbf{23.79}$ & $\textbf{27.39}$ & $25.44$ & $\underline{27.65}$ & $\underline{27.06}$ & $\textbf{30.27}$ & $26.05$ & $22.52$ & $\underline{25.89}$ \\
        &                  & \textbf{Ours}   & $25.99$ & $21.54$ & $22.95$ & $26.83$ & $\textbf{26.25}$ & $\textbf{28.48}$ & $\textbf{27.45}$ & $29.78$ & $\underline{27.00}$ & $\textbf{24.50}$ & $\textbf{26.08}$ \\
    \cmidrule(lr){3-14}
     & SSIM $\uparrow$ & 3DGS            & $0.889$ & $0.787$ & $0.812$ & $0.914$ & $0.863$ & $0.922$ & $0.896$ & $0.892$ & $0.867$ & $0.768$ & $0.861$ \\
        &                  & ThermalGaussian & $0.886$ & $0.835$ & $0.862$ & $0.919$ & $\underline{0.874}$ & $0.922$ & $0.888$ & $0.896$ & $0.883$ & $0.850$ & $0.882$ \\
        &                  & MS-Splattingv2  & $0.876$ & $0.803$ & $0.855$ & $0.900$ & $0.871$ & $0.927$ & $0.888$ & $0.890$ & $\underline{0.903}$ & $0.853$ & $0.877$ \\
        &                  & MMOne           & $\textbf{0.894}$ & $\textbf{0.840}$ & $\textbf{0.874}$ & $\underline{0.926}$ & $0.870$ & $\underline{0.933}$ & $\underline{0.902}$ & $\underline{0.906}$ & $0.895$ & $\underline{0.861}$ & $\underline{0.890}$ \\
        &                  & \textbf{Ours}   & $\underline{0.890}$ & $\underline{0.839}$ & $\underline{0.865}$ & $\textbf{0.928}$ & $\textbf{0.889}$ & $\textbf{0.941}$ & $\textbf{0.909}$ & $\textbf{0.910}$ & $\textbf{0.912}$ & $\textbf{0.889}$ & $\textbf{0.897}$ \\
    \cmidrule(lr){3-14}
     & LPIPS $\downarrow$ & 3DGS            & $0.127$ & $0.259$ & $0.307$ & $0.209$ & $0.142$ & $0.126$ & $0.185$ & $0.091$ & $0.227$ & $0.378$ & $0.205$ \\
        &                  & ThermalGaussian & $0.129$ & $0.210$ & $0.203$ & $0.198$ & $0.136$ & $0.124$ & $0.177$ & $0.091$ & $0.181$ & $0.248$ & $0.170$ \\
        &                  & MS-Splattingv2  & $\textbf{0.092}$ & $\underline{0.164}$ & $\textbf{0.148}$ & $\underline{0.133}$ & $\underline{0.107}$ & $\underline{0.073}$ & $\underline{0.103}$ & $\underline{0.064}$ & $\textbf{0.075}$ & $\underline{0.177}$ & $\underline{0.114}$ \\
        &                  & MMOne           & $0.125$ & $0.194$ & $0.201$ & $0.213$ & $0.142$ & $0.127$ & $0.198$ & $0.083$ & $0.205$ & $0.272$ & $0.176$ \\
        &                  & \textbf{Ours}   & $\underline{0.100}$ & $\textbf{0.149}$ & $\underline{0.149}$ & $\textbf{0.096}$ & $\textbf{0.094}$ & $\textbf{0.059}$ & $\textbf{0.085}$ & $\textbf{0.057}$ & $\underline{0.075}$ & $\textbf{0.139}$ & $\textbf{0.101}$ \\
\bottomrule
\end{tabular}
}
\end{table*}

\subsection{Ablation}
To verify the contribution of each design element, we conduct ablation experiments as summarized in Tab.~\ref{tab:ablation}. First, the comparison between our full model and the ``MLP-based'' variant demonstrates the advantage of our feature-guided modulation over a standard decoding structure, as the former better leverages spatial-aware features for high-quality appearance synthesis. Second, removing the geometric decoupling mechanism (``w.o. geo. decoup.'') leads to a consistent performance decline in both modalities, confirming that isolating physical geometry from modality-specific radiation is essential for robust RGBT scene modeling. Finally, the exclusion of latent constraints (``w.o. fea(rgb/th)'') results in degradation of perceptual details, which indicates that the feature-level reconstruction loss $\mathcal{L}_{rec}^{feat}$ is crucial for encouraging the model to capture fundamental structural and intensity distributions. Collectively, these results validate that the synergy of the proposed modulation mechanism, geometric decoupling, and latent supervision ensures optimal RGBT reconstruction.

\begin{table}[ht]
\centering
\caption{Ablation Study. We conducted ablation experiments on different modules of our pipeline.}
\label{tab:ablation}
\setlength{\tabcolsep}{3pt}
\begin{tabular}{lccc|ccc}
\toprule
 & \multicolumn{3}{c|}{RGB Modality} & \multicolumn{3}{c}{Thermal Modality} \\
Method Variant & PSNR $\uparrow$ & SSIM $\uparrow$ & LPIPS $\downarrow$ & PSNR $\uparrow$ & SSIM $\uparrow$ & LPIPS $\downarrow$ \\
\midrule
MLP-based                 & $\underline{25.14}$ & $\underline{0.869}$ & $\underline{0.154}$ & $25.88$ & $0.895$ & $\underline{0.104}$ \\
w.o. geo. decoup.    & $25.07$ & $0.868$ & $0.159$ & $25.82$ & $0.892$ & $0.106$ \\
w.o. hybrid rgb           & $25.07$ & $0.867$ & $0.157$ & $25.86$ & $0.893$ & $0.105$ \\
w.o. fea(th)              & $24.98$ & $0.868$ & $0.155$ & $25.77$ & $0.894$ & $0.107$ \\
w.o. fea(rgb)             & $24.88$ & $0.858$ & $0.190$ & $\underline{25.93}$ & $\textbf{0.897}$ & $0.104$ \\
\midrule
\textbf{Ours}                      & $\textbf{25.23}$ & $\textbf{0.871}$ & $\textbf{0.153}$ & $\textbf{26.08}$ & $\underline{0.897}$ & $\textbf{0.101}$ \\
\bottomrule
\end{tabular}
\end{table}

\section{conclusion}
In this paper, we present \ourpaper, a novel cross-modal 3D Gaussian Splatting framework designed for high-fidelity RGBT scene reconstruction. To effectively bridge the gap between visible and thermal modalities, we introduce a Spectrum-Aware Adaptive Modulation mechanism that leverages thermal structural priors to guide visible texture synthesis. Furthermore, to address the inherent geometric inconsistencies caused by disparate physical sensing properties, we propose a Modality-Adaptive Geometric Decoupling scheme, which enables the model to accurately represent independent spectral characteristics without compromising spatial alignment. Extensive experiments on the RGBT-Scenes dataset demonstrate that our approach achieves state-of-the-art performance in both rendering quality and structural accuracy. By integrating explicit geometric representations with implicit neural feature modulation, \ourpaper provides a robust and efficient solution for multi-spectral scene understanding in visually degraded environments.

\textbf{limitations.}
Despite the promising results, \ourpaper has certain limitations that offer directions for future research. First, the current geometric decoupling scheme primarily focuses on the thermal branch; however, in scenarios with extreme glass reflections or high-transparency surfaces, more complex multi-modal interactions might be required to fully resolve depth ambiguities. Second, the use of latent feature modulation introduces additional memory overhead during the neural decoding phase compared to vanilla 3DGS. 
\szq{Third, unlike MMOne~\cite{mmone}, our current framework does not support language field reconstruction, as we focus purely on RGB and thermal modalities.}
Future work will explore more lightweight modulation architectures and investigate the potential of extending this framework to other spectral domains, such as near-infrared or hyperspectral data, to further enhance its versatility and robustness in all-weather environmental perception. \szq{Additionally, we plan to incorporate language field reconstruction and explore broader downstream applications.}

\section*{Acknowledgments}
This paper is supported in part by the National Natural Science Foundation of China (Grant No. 62402274) and the Start-up Funding of Fuzhou University (Grant No. XRC-25164) to Zhaoqi Su; in part by the Education and Scientific Research Project for Middle-aged and Young Teachers of Fujian Province, China (Grant No. JZ250004) to Zhipeng Su; in part by the Special Fund for Promoting High-Quality Development of Marine and Fishery Industries in Fujian Province (Grant No. FJHYF-L-2025-07-005) to Xiaoqiang Lu. 
During the preparation of this work the authors used Gemini in order to improve the language and readability of the manuscript during the writing process. After using this tool, the authors reviewed and edited the content as needed and take full responsibility for the content of the published article.
\section*{Acknowledgments}
This paper is supported in part by the National Natural Science Foundation of China (Grant No. 62402274) and the Start-up Funding of Fuzhou University (Grant No. XRC-25164) to Zhaoqi Su; in part by the Education and Scientific Research Project for Middle-aged and Young Teachers of Fujian Province, China (Grant No. JZ250004) to Zhipeng Su; in part by the Special Fund for Promoting High-Quality Development of Marine and Fishery Industries in Fujian Province (Grant No. FJHYF-L-2025-07-005) to Xiaoqiang Lu. The authors would like to acknowledge the use of Gemini to improve the language and readability of the manuscript during the writing process.

 


\bibliographystyle{IEEEtran}
\bibliography{refs}

\newpage



\vfill

\end{document}